\documentclass[letterpaper, 10 pt, conference]{ieeeconf}  

\IEEEoverridecommandlockouts                              

\usepackage{graphics} 
\usepackage{epsfig} 
\usepackage{mathptmx} 
\usepackage{times} 
\usepackage{amsmath} 
\usepackage{amssymb}  
\usepackage{censor}
\usepackage{balance}

\title{\LARGE \bf
DreamToNav: Generalizable Navigation for Robots via Generative Video Planning
}

\author{Valerii Serpiva, Jeffrin Sam, Chidera Simon, Hajira Amjad, Iana~Zhura, Artem Lykov, and Dzmitry Tsetserukou
\thanks{The authors are with the Intelligent Space Robotics Laboratory, Skolkovo Institute of Science and Technology Moscow, Bolshoy Boulevard 30, bld. 1, 121205, Moscow, Russia.
\tt \{Valerii.Serpiva, Jeffrin.Sam, Chidera.Agbasiere, Hajira.Amjad, Iana.Zhura, Artem.Lykov, D.Tsetserukou\}@skoltech.ru}}

\begin{document}

\maketitle
\thispagestyle{empty}
\pagestyle{empty}

\begin{abstract}


We present DreamToNav, a novel autonomous robot framework that uses generative video models to enable intuitive, human-in-the-loop control. Instead of relying on rigid waypoint navigation, users provide natural language prompts (e.g. ``Follow the person carefully''), which the system translates into executable motion. Our pipeline first employs Qwen 2.5-VL-7B-Instruct to refine vague user instructions into precise visual descriptions. These descriptions condition NVIDIA Cosmos 2.5, a state-of-the-art video foundation model, to synthesize a physically consistent video sequence of the robot performing the task. From this synthetic video, we extract a valid kinematic path using visual pose estimation, robot detection and trajectory recovery. By treating video generation as a planning engine, DreamToNav allows robots to visually ``dream'' complex behaviors before executing them, providing a unified framework for obstacle avoidance and goal-directed navigation without task-specific engineering. We evaluate the approach on both a wheeled mobile robot and a quadruped robot in indoor navigation tasks. DreamToNav achieves a success rate of 76.7\%, with final goal errors typically within 0.05-0.10\,m and trajectory tracking errors below 0.15\,m. These results demonstrate that trajectories extracted from generative video predictions can be reliably executed on physical robots across different locomotion platforms.

\end{abstract}

\section{Introduction}

Autonomous navigation of robots in human-populated spaces requires more than collision-free path planning; it requires a semantic understanding of the environment and the ability to interpret high-level user intent. A command such as ``Follow that person politely'' entails complex spatial reasoning about interpersonal distance, timing, and socially appropriate motion behaviors that are exceedingly difficult to encode through hand-crafted cost functions or rule-based planners~\cite{10.3389/frobt.2021.721317}. While classical approaches decompose navigation into mapping, localization, and trajectory optimization, they often fail to capture the nuanced semantics embedded in natural language instructions.

Recent advances in foundation models have opened new avenues for improving human robot interaction by bridging the gap between natural language instructions and robotic navigation capabilities~\cite{yakolli2025surveyimprovinghumanrobot}. Large Vision-Language Models (LVLMs)~\cite{lou-etal-2024-large} demonstrate remarkable scene understanding and instruction following, while generative video models have shown the capacity to synthesize physically plausible future sequences. A key insight motivating this work is that if a generative model can produce a realistic video of a robot performing a task, it has effectively solved the planning problem~\cite{Hu_2023_Gaia1}. Prior work at cognitive Visual Language Action (VLA) models for drone navigation~\cite{wu2025vlaanefficientonboardvisionlanguageaction}, while FlightDiffusion~\cite{Serpiva_2025_FlightDiffusion} demonstrated diffusion-based video generation as a viable planning engine for unmanned aerial vehicles. Concurrently, NVIDIA's Alpamayo-R1~\cite{Wang_2025_Alpamayo} showed that coupling vision-language reasoning with action prediction via Chain-of-Causation reasoning significantly improves decision-making in safety-critical autonomous driving scenarios, further motivating the integration of world foundation models with language-guided planning.

\begin{figure}[t]
    \centering
    \includegraphics[width=0.5\textwidth]{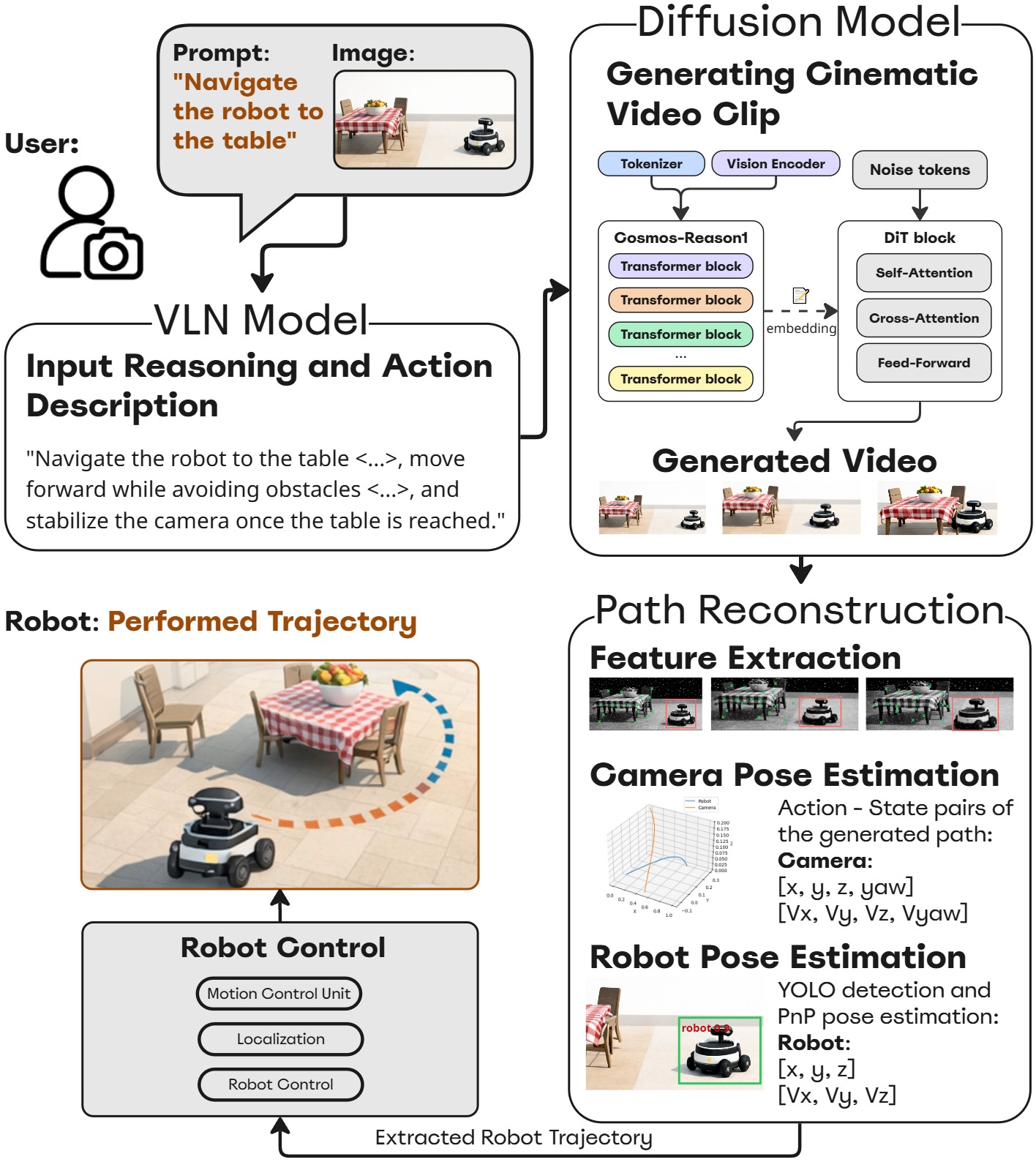}
    \caption{Pipeline for trajectory extraction from a single image and text prompt. A user provides an input image of a robot and a textual prompt describing the desired motion. The image prompt pair is processed by a VLM and the COSMOS video generation module to synthesize a plausible motion sequence. Visual odometry and pose estimation are then applied to the generated video to recover robot poses and track visual features over time. The final output is the estimated trajectory of the robot derived solely from the initial image and text prompt.}
    \label{fig:teaser}
\end{figure}

In this work, we demonstrate the applicability of the video-as-planning paradigm across different robotic platforms, including ground and legged robots. DreamToNav enables robots to “dream” their actions before executing them. Furthermore, it introduces a new human-robot interaction paradigm where users specify navigation tasks using only a scene image and a natural language prompt, allowing intuitive control without explicit trajectory or waypoint definition. The system combines the reasoning capabilities of Qwen~2.5-VL~\cite{Bai_2023_QwenVL} for prompt refinement with the physics-aware video generation of NVIDIA Cosmos 2.5~\cite{NVIDIA_2025_Cosmos} to synthesize visually consistent future frame sequences. A trajectory is then extracted from the generated video using object detection, pose estimation, and visual odometry, and subsequently executed on the physical robot. By decoupling high-level reasoning from physical prediction, DreamToNav enables robots to execute diverse natural language commands including obstacle avoidance, person following, and exploratory navigation without task-specific engineering or reward shaping.


\section{Related Work}

\subsection{Generative World Models for Robotic Control}

The paradigm of using generative models as implicit planners has gained substantial traction across robotics and autonomous driving. Du et al.\ introduced UniPi~\cite{Du_2023_UniPi}, demonstrating that a text-conditioned video diffusion model can serve as a universal policy for robotic manipulation by generating future visual plans from which actions are inferred via inverse dynamics. In the autonomous driving domain, GAIA-1~\cite{Hu_2023_Gaia1} proposed a generative world model that conditions on video, text, and action inputs to predict realistic driving scenarios, effectively functioning as a high-fidelity simulator. DriveDreamer~\cite{Wang_2024_DriveDreamer} extended this concept by using diffusion models to generate controllable driving videos aligned with traffic constraints, enabling both future prediction and policy extraction. Similarly, Panacea~\cite{Wen_2024_Panacea} proposed panoramic and controllable video generation for simulating complex road scenarios.

More recently, NVIDIA's Alpamayo-R1~\cite{Wang_2025_Alpamayo} bridged reasoning and action prediction by combining a VLA architecture with Chain-of-Causation reasoning and a diffusion-based trajectory decoder, achieving significant improvements in long-tail driving scenarios. Their use of Cosmos-Reason, a Visual Language Model (VLM) pre-trained for Physical AI applications, resonates strongly with the approach adopted in this work of leveraging the Cosmos world foundation model for physics-aware video generation.

DreamToNav extends the ``video-as-planning'' paradigm to social robot navigation. Unlike driving-centric methods that operate in structured road environments, the proposed system targets unstructured indoor and outdoor spaces where semantic understanding of social cues is critical. 

\subsection{Diffusion-Based Navigation}

Diffusion models have emerged as a powerful alternative to reinforcement learning for multimodal navigation tasks. NoMaD~\cite{Sferrazza_2024_NoMaD} introduced a goal-masked diffusion policy that generates feasible trajectories directly from visual observations, eliminating the need for explicit maps. Trajectory Diffusion~\cite{Li_2024_TrajectoryDiffusion} generates waypoints for object-goal navigation in latent space, while NaviDiffuser~\cite{NaviDiffuser_2025} addresses multi-objective navigation by balancing competing constraints through diffusion-based planning.

However, the majority of these approaches operate in learned latent spaces, which limits interpretability. By contrast, methods that generate visible video predictions~\cite{Zhang_2025_VILP} offer a distinct advantage: a human operator can inspect the robot's intended plan before execution. DreamToNav adopts this transparent planning philosophy, generating explicit third-person video predictions that serve as both a planning artifact and a human-readable explanation of the robot's intended behavior.

\subsection{Vision-Language Models for Social Navigation}

Social navigation requires robots to interpret visual and linguistic cues that govern human spatial behavior. Classical methods, such as the representation learning approach of Karnychev and Ferrer~\cite{Karnychev_2022_SocialNav}, learn socially compliant path features from demonstrations but lack the flexibility to handle open-ended language instructions. The integration of LVLMs offers a path forward: DriveVLM~\cite{Tian_2024_DriveVLM} demonstrated that chain-of-thought reasoning over visual scenes enables more nuanced planning decisions for ground vehicles, while Genie~\cite{Bruce_2024_Genie} showed that generative interactive environments can capture complex physical dynamics.

DreamToNav leverages Qwen~2.5-VL~\cite{Bai_2023_QwenVL} as a reasoning front-end that transforms ambiguous user prompts into structured visual descriptions suitable for video generation. This decoupled architecture separating semantic reasoning VLM from physical world simulation (Cosmos~2.5) allows the system to handle nuanced instructions such as ``keep a safe distance'' or ``approach from the left'' without requiring explicit cost-map tuning or task-specific training data.
\section{Methodology}

The DreamToNav pipeline, which consists of three stages: Prompt Refinement, Video Generation, and Trajectory Extraction, is illustrated in the following system architecture diagram~\ref{fig:teaser}.


\subsection{Prompt Refinement via Qwen 2.5-VL}

Raw user prompts (e.g., ``go there'') are often too spatially underspecified 
for reliable video generation. We employ \textbf{Qwen 2.5-VL-7B-Instruct}~\cite{Bai_2023_QwenVL} as a multimodal reasoning agent to bridge this 
semantic gap before any video synthesis takes place.

Given the current camera frame $I_0$ and the raw user instruction 
$p_{\text{raw}}$, Qwen 2.5-VL executes a three-stage reasoning pass. 
\textbf{(1) Scene Grounding:} The model parses $I_0$ to identify salient 
objects, their spatial relationships, and potential obstacles in the 
environment. \textbf{(2) Reference Resolution:} Ambiguous linguistic 
expressions (e.g., ``there,'' ``it,'' ``that obstacle'') are resolved by 
anchoring them to specific entities detected in the scene. \textbf{(3) 
Motion Decomposition:} The operator's high-level intent is decomposed into 
motion primitives expressible as visual descriptions, including direction, 
approximate speed, and social constraints such as safe interpersonal 
distance. The output of this reasoning chain is a structured natural 
language description grounded in metric and directional terms, for example: 
\textit{``The camera moves forward at 1\,m/s, smoothly curving left by 
30$^\circ$ to avoid the chair, then re-centering toward the hallway exit.''}

This scene-aware, grounded prompt maximizes the physical plausibility and 
prompt adherence of the downstream video generation model, acting as a 
zero-shot spatial translator between ambiguous human intent and precise 
visual specifications, without requiring any task-specific fine-tuning.

\subsection{Video Generation via Cosmos 2.5}

Given the refined prompt $\hat{p}$ and the initial frame $I_0$, we 
synthesize a ``visual plan'' using \textbf{NVIDIA Cosmos 2.5}~\cite{NVIDIA_2025_Cosmos}, a world foundation model pre-trained on 
large-scale physical interaction data to predict physically plausible 
future states. Unlike artistic diffusion models optimized primarily for 
perceptual aesthetics~\cite{Bruce_2024_Genie}, Cosmos 2.5 is trained to 
model realistic kinematic constraints, object permanence, and environment 
dynamics, making it directly applicable to robotic trajectory planning.

The video synthesis proceeds as a conditional latent denoising process. 
Starting from Gaussian noise $\mathbf{z}_T \sim \mathcal{N}(\mathbf{0}, 
\mathbf{I})$ in a compressed spatio-temporal latent space, Cosmos 2.5 
iteratively denoises the latent tensor over $T$ diffusion timesteps 
according to:
\begin{equation}
    \mathbf{z}_{t-1} = f_\theta\!\left(\mathbf{z}_t,\, t,\, 
    \phi(\hat{p}),\, \psi(I_0)\right),
\end{equation}
where $f_\theta$ is the learned denoising network, $\phi(\hat{p})$ denotes 
the text-conditioned prompt embeddings encoding motion intent (direction, 
speed, social constraints), and $\psi(I_0)$ denotes the visual context 
embeddings encoding the observed scene geometry and object layout from the 
initial frame. This dual conditioning ensures the generated sequence 
simultaneously respects the operator's semantic intent and the physical 
structure of the real environment. The denoised latent is decoded into a 
sequence of $N$ RGB frames:
\begin{equation}
    V_{\mathrm{syn}} = \{I_k\}_{k=1}^{N} = \mathcal{D}(\mathbf{z}_0),
\end{equation}
where $\mathcal{D}$ is the latent-to-pixel decoder.

To provide richer spatial information for the downstream trajectory extraction stage, we generate a synthetic \textbf{Third-Person View (TPV)} from the same conditioning inputs by modifying the camera perspective tokens. The TPV is generated by conditioning the video model on an elevated external camera viewpoint~\cite{Wen_2024_Panacea}, producing frames that depict the robot’s motion from a global perspective. This view provides clear spatial context of the robot relative to surrounding obstacles and navigation targets, which facilitates robust pose estimation and trajectory extraction. By observing the robot from an external viewpoint, the TPV reduces ambiguities in robot localization and improves the stability of the recovered motion trajectory in the subsequent planning pipeline.

\subsection{Robot Detection Dataset and Training}

\begin{figure}[t]
  \centering
  \includegraphics[width=0.5\textwidth]{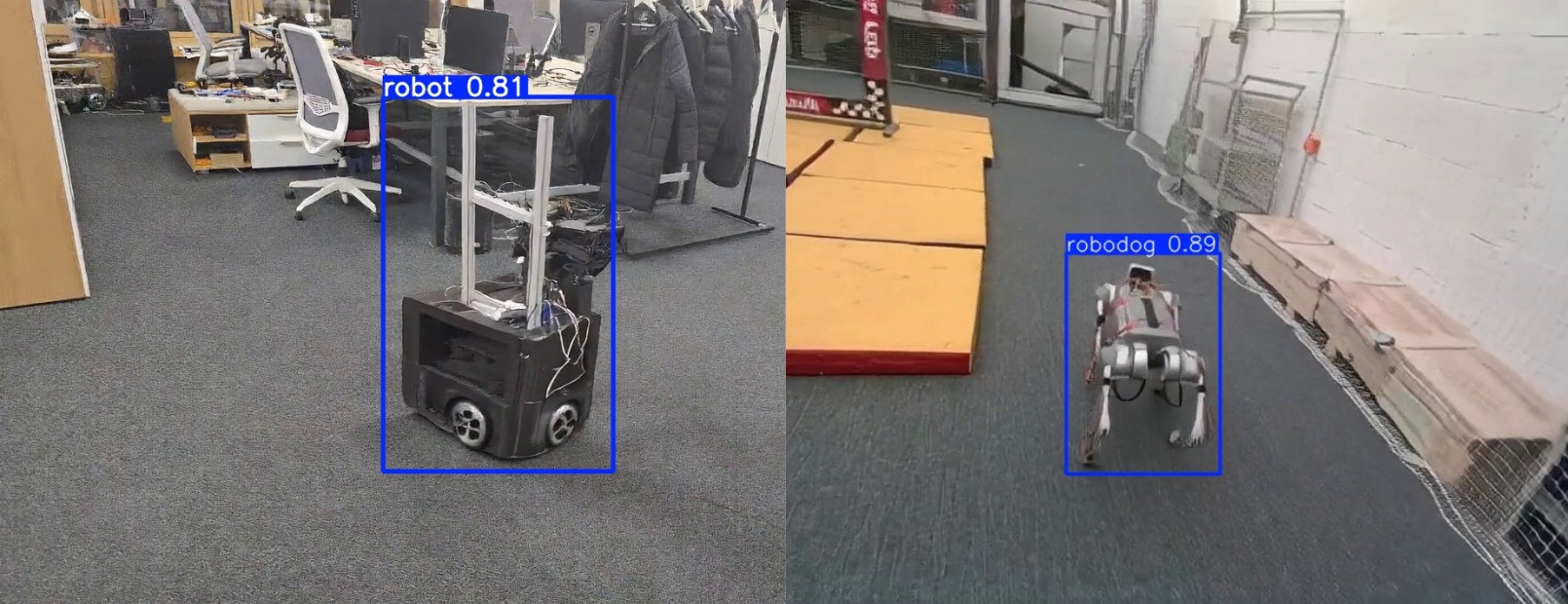}
  \caption{Robot detection results using the trained YOLO11n model. The detector identifies both a wheeled UGV platform and a quadruped robot dog on diffused and real frames.}
  \label{fig:detection}
\end{figure}

To enable reliable detection of robotic platforms in generated and real-world scenes, we collected a small dataset containing images of both Unmanned Ground Vehicles (UGVs) and quadruped (robot dog) platforms. The dataset consists of real images recorded during robot experiments as well as synthetically generated images obtained using diffusion-based generative models.

Real images were captured in indoor laboratory environments during robot navigation experiments. In addition, diffusion-based image generation was used to augment the dataset with diverse viewpoints, lighting conditions, and backgrounds. Some of the generated frames include partially hallucinated robot appearances, which increases visual diversity and improves model robustness to imperfect visual inputs.

All images were manually annotated with bounding boxes corresponding to the robot instances. The final dataset includes examples of both wheeled UGV platforms and legged robots.

For object detection, we trained a lightweight YOLO11n model due to its favorable trade-off between detection accuracy and computational efficiency, which is important for real-time robotic applications. The model was trained on the collected dataset to detect robotic platforms in both real and generated frames used within the proposed video-based planning pipeline.

\subsection{Trajectory Extraction \& Execution}

Let the synthesized video be represented as a sequence of RGB frames
\begin{equation}
    \mathcal{I} = \{ I_k \}_{k=1}^{N},
\end{equation}
where $k$ indexes time (frame index).

We employ ORB-SLAM3~\cite{ORBSLAM3_TRO} to estimate the pose of the virtual camera for each frame. The pose is modeled as a rigid transformation in $\mathrm{SE}(3)$ from the world frame $\mathcal{F}_w$ to the camera frame $\mathcal{F}_c$:
\begin{equation}
    \mathbf{T}_{w c,k} 
    =
    \begin{bmatrix}
        \mathbf{R}_{w c,k} & \mathbf{t}_{w c,k} \\
        \mathbf{0}^\top & 1
    \end{bmatrix} 
    \in \mathrm{SE}(3),
\end{equation}
where $\mathbf{R}_{w c,k} \in \mathrm{SO}(3)$ is the rotation matrix and $\mathbf{t}_{w c,k} \in \mathbb{R}^3$ is the translation vector at frame $k$. ORB-SLAM3 provides an estimate
\begin{equation}
    \hat{\mathbf{T}}_{w c,k} = \bigl( \hat{\mathbf{R}}_{w c,k}, \hat{\mathbf{t}}_{w c,k} \bigr)
\end{equation}
for each frame $k$.

For each frame $I_k$, we detect the robot using a YOLOv11n object detection model. The detector outputs a bounding box
\begin{equation}
    \mathbf{b}_k = (u_k, v_k, w_k, h_k),
\end{equation}
where $(u_k, v_k)$ denotes the center of the bounding box in pixel coordinates and $(w_k, h_k)$ its width and height. This bounding box is used as an input cue for subsequent 3D pose estimation.

We denote the 2D image points associated with the robot as
\begin{equation}
    \mathbf{p}_{i,k} = 
    \begin{bmatrix}
        u_{i,k} \\[2pt]
        v_{i,k}
    \end{bmatrix},
    \quad i = 1, \dots, M,
\end{equation}
where $M$ is the number of feature points (e.g., corners or keypoints) extracted from the bounding box region.

\begin{figure}[t]
  \centering
  \includegraphics[width=0.5\textwidth]{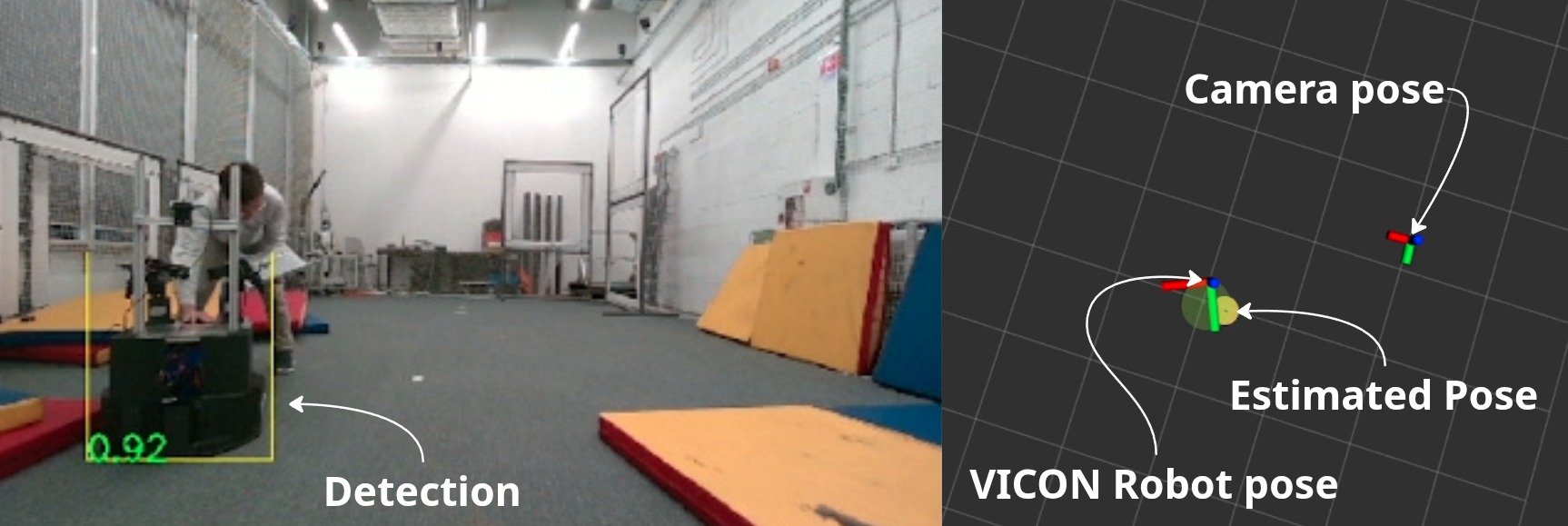}
  \caption{Robot pose estimation pipeline. Left: robot detection in the camera image. Right: estimated robot trajectory obtained from visual odometry and pose estimation compared with the ground-truth robot pose recorded by the VICON motion capture system.}
  \label{fig:pose_est}
\end{figure}

We assume the robot has known physical dimensions, enabling us to define a simple 3D model in the robot frame $\mathcal{F}_r$. Let
\begin{equation}
    \mathbf{P}_i^r =
    \begin{bmatrix}
        X_i^r \\[2pt]
        Y_i^r \\[2pt]
        Z_i^r
    \end{bmatrix},
    \quad i = 1, \dots, M,
\end{equation}
be the 3D coordinates of model points in $\mathcal{F}_r$. The camera intrinsics are given by
\begin{equation}
    \mathbf{K} =
    \begin{bmatrix}
        f_x & 0   & c_x \\
        0   & f_y & c_y \\
        0   & 0   & 1
    \end{bmatrix}.
\end{equation}

The projection of a 3D point $\mathbf{P}_i^r$ into the image plane at frame $k$ is modeled by
\begin{equation}
    s_{i,k}
    \begin{bmatrix}
        u_{i,k} \\[2pt]
        v_{i,k} \\[2pt]
        1
    \end{bmatrix}
    = 
    \mathbf{K}
    \Bigl[
        \mathbf{R}_{c r,k} \ \big|\ \mathbf{t}_{c r,k}
    \Bigr]
    \begin{bmatrix}
        \mathbf{P}_i^r \\[2pt]
        1
    \end{bmatrix},
\end{equation}
where $\mathbf{R}_{c r,k} \in \mathrm{SO}(3)$ and $\mathbf{t}_{c r,k} \in \mathbb{R}^3$ denote the rotation and translation from the robot frame $\mathcal{F}_r$ to the camera frame $\mathcal{F}_c$ at time $k$, and $s_{i,k}$ is a depth-dependent scale.

\begin{figure}[t]
  \centering
  \includegraphics[width=0.5\textwidth]{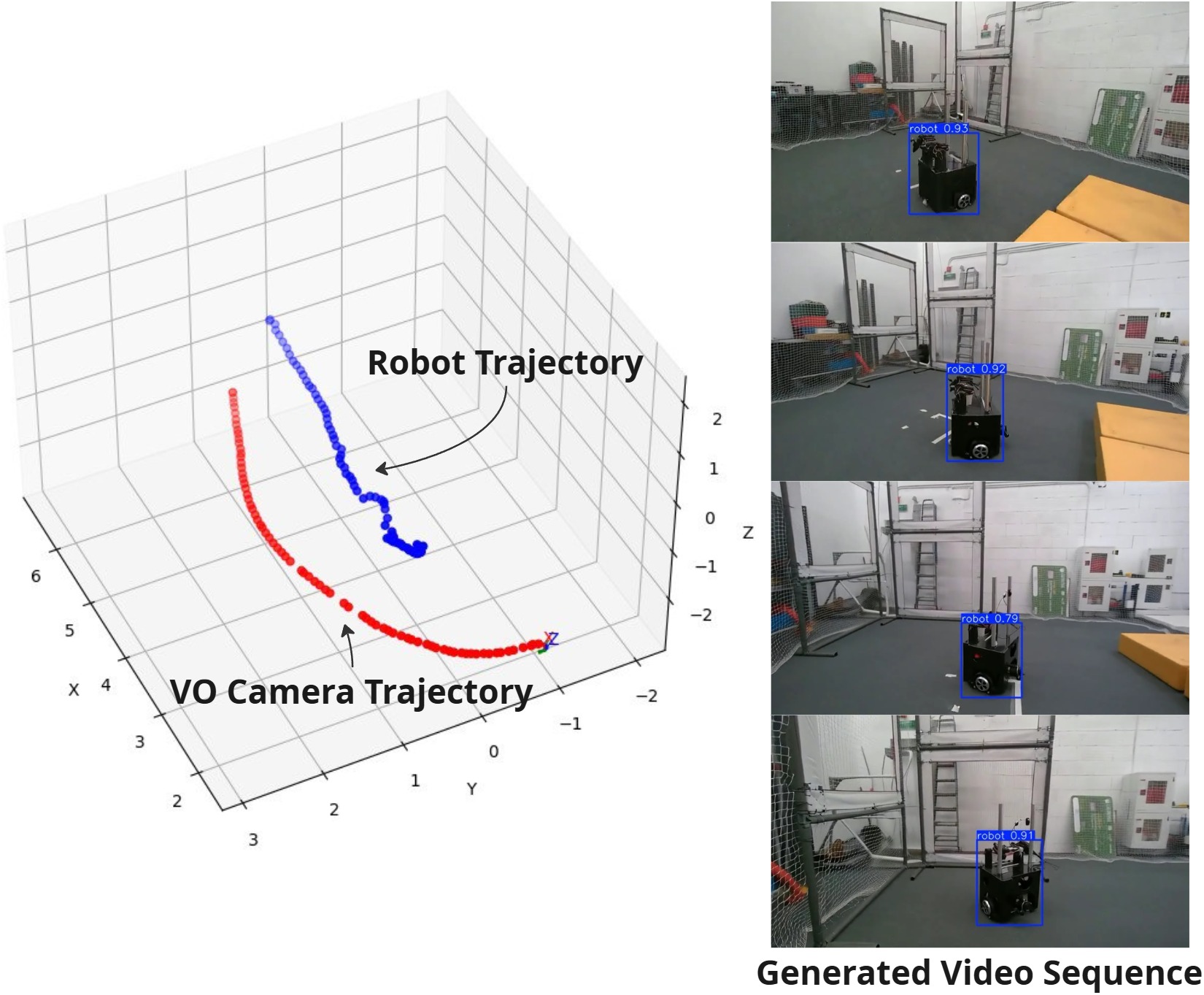}
  \caption{Trajectory extraction from generated video frames. The robot is detected in each generated frame (right), and its pose is estimated using PnP to recover the robot trajectory (blue). The corresponding camera trajectory obtained from visual odometry is shown in red.}
  \label{fig:vi_pois}
\end{figure}

We use the IPPE-based PnP~\cite{Collins2014} algorithm to estimate $(\mathbf{R}_{c r,k}, \mathbf{t}_{c r,k})$ by minimizing the reprojection error:
\begin{equation}
    (\hat{\mathbf{R}}_{c r,k}, \hat{\mathbf{t}}_{c r,k})
    = 
    \arg\min_{\mathbf{R}, \mathbf{t}}
    \sum_{i=1}^{M}
    \left\|
        \mathbf{p}_{i,k} - \pi \bigl( \mathbf{K}, \mathbf{R}, \mathbf{t}, \mathbf{P}_i^r \bigr)
    \right\|^2,
\end{equation}
where $\pi(\cdot)$ denotes the perspective projection function.

The resulting robot position in the camera frame is
\begin{equation}
    \mathbf{p}_{r,k}^c = \hat{\mathbf{t}}_{c r,k}
    = \begin{bmatrix} x_{r,k}^c & y_{r,k}^c & z_{r,k}^c \end{bmatrix}^\top.
\end{equation}
In this work, we focus on estimating the robot position $(x,y,z)$.

To reduce noise and enforce temporal consistency, we apply an Extended Kalman Filter to the sequence of robot positions in the camera frame. We define the state vector at time $k$ as
\begin{equation}
    \mathbf{x}_k =
    \begin{bmatrix}
        \mathbf{p}_{r,k}^c & \mathbf{v}_{r,k}^c
    \end{bmatrix}^\top
    =
    \begin{bmatrix}
        x_{r,k}^c & y_{r,k}^c & z_{r,k}^c &
        \dot{x}_{r,k}^c & \dot{y}_{r,k}^c & \dot{z}_{r,k}^c
    \end{bmatrix}^\top
    \in \mathbb{R}^6.
\end{equation}

where $\mathbf{v}_{r,k}^c$ is the robot velocity in the camera frame.

Assuming a constant-velocity motion model with sampling interval $\Delta t_k$, the process model is
\begin{equation}
    \mathbf{x}_{k+1} = \mathbf{F}_k \mathbf{x}_k + \mathbf{w}_k,
\end{equation}
with
\begin{equation}
    \mathbf{F}_k =
    \begin{bmatrix}
        \mathbf{I}_3 & \Delta t_k \mathbf{I}_3 \\
        \mathbf{0}_3 & \mathbf{I}_3
    \end{bmatrix},
\end{equation}
where $\mathbf{I}_3$ is the $3 \times 3$ identity matrix, $\mathbf{0}_3$ is the $3 \times 3$ zero matrix, and $\mathbf{w}_k \sim \mathcal{N}(\mathbf{0}, \mathbf{Q}_k)$ is zero-mean process noise.

The measurement at time $k$ is given by the noisy position estimate from PnP:
\begin{equation}
    \mathbf{z}_k = \mathbf{p}_{r,k}^c + \mathbf{v}_k,
\end{equation}
with $\mathbf{v}_k \sim \mathcal{N}(\mathbf{0}, \mathbf{R}_k)$ the measurement noise. The measurement model can be written as
\begin{equation}
    \mathbf{z}_k = \mathbf{H} \mathbf{x}_k + \mathbf{v}_k,
\end{equation}
where
\begin{equation}
    \mathbf{H} =
    \begin{bmatrix}
        \mathbf{I}_3 & \mathbf{0}_3
    \end{bmatrix}.
\end{equation}

The filtered robot position in the camera frame is extracted as
\begin{equation}
    \tilde{\mathbf{p}}_{r,k}^c =
    \begin{bmatrix}
        \tilde{x}_{r,k}^c &
        \tilde{y}_{r,k}^c &
        \tilde{z}_{r,k}^c
    \end{bmatrix}^\top
    = \left[ \hat{\mathbf{x}}_{k|k} \right]_{1:3}.
\end{equation}

Finally, we transform the filtered robot positions from the camera frame to the world frame using the estimated virtual camera pose. For each time $k$,
\begin{equation}
    \tilde{\mathbf{p}}_{r,k}^w
    =
    \hat{\mathbf{R}}_{w c,k} \, \tilde{\mathbf{p}}_{r,k}^c
    + \hat{\mathbf{t}}_{w c,k},
\end{equation}
where $\tilde{\mathbf{p}}_{r,k}^w \in \mathbb{R}^3$ is the robot position in the world frame.

The resulting robot trajectory is given by the sequence
\begin{equation}
    \mathcal{T}_r 
    = 
    \left\{
        \tilde{\mathbf{p}}_{r,1}^w,
        \tilde{\mathbf{p}}_{r,2}^w,
        \dots,
        \tilde{\mathbf{p}}_{r,N}^w
    \right\}.
\end{equation}
This trajectory is then used as the reference path for the robot to execute in the real environment.

To execute the estimated path, the 3D robot trajectory is projected onto the ground plane, yielding a 2D reference curve
\begin{equation}
    \mathcal{T}_r^{2D} = \{ (x_k, y_k) \}_{k=1}^{N}.
\end{equation}

The final robot trajectory recovered from the generated video sequence is presented in Fig.~\ref{fig:vi_pois}.


\section{Experimental Evaluation}
\label{sec:experiments}

We evaluate \textit{DreamToNav} on two robotic platforms: a wheeled UGV and a quadruped robot. The objective of the experiments is to assess whether trajectories extracted from generative video predictions can be executed reliably on real robots in cluttered indoor environments. In all experiments, the user first captures an image of the scene with a robot and then provides a natural-language command, such as \textit{``navigate to the blue object while avoiding collisions''}. The system then generates a future video of the robot executing the task, estimates the robot poses across the generated frames, converts these poses into a navigation trajectory, and finally executes the trajectory on the physical robot.

\subsection{Experimental Pipeline}

\begin{figure*}[t]
  \centering
  \includegraphics[width=0.85\textwidth]{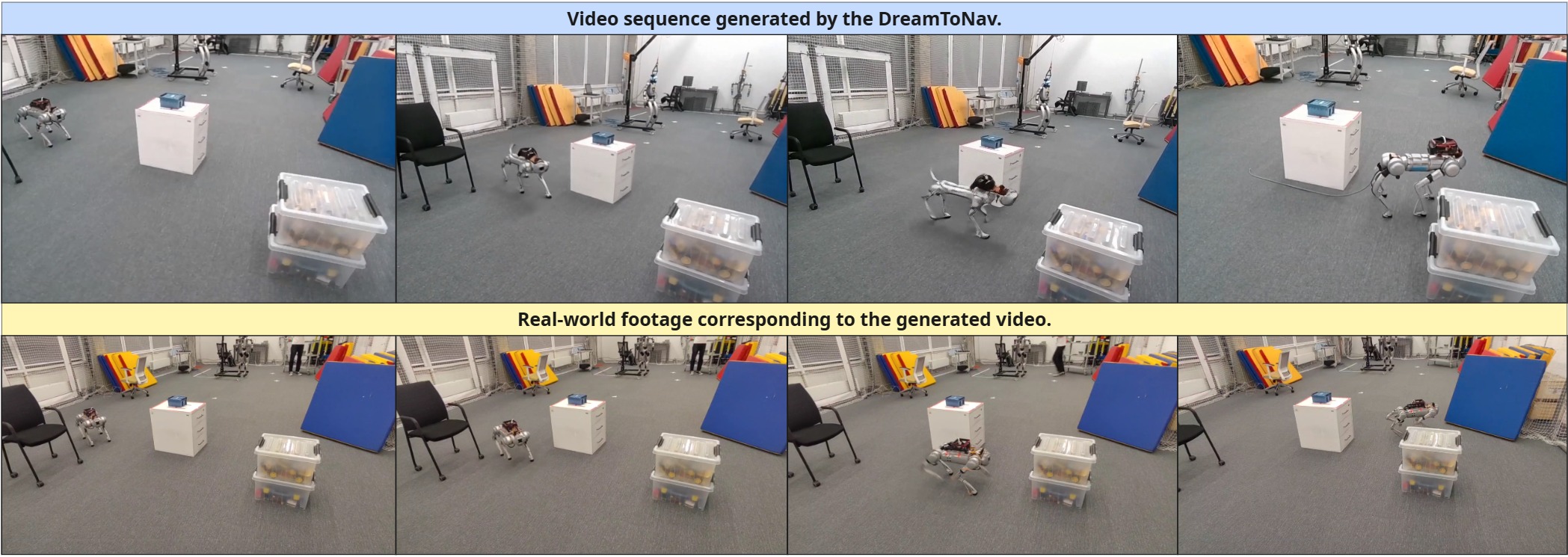}
  \caption{Comparison between the video sequence generated by DreamToNav (top row) and the corresponding real-world execution (bottom row). Given a scene image and a natural language instruction, the system generates a future video predicting the robot's motion while avoiding obstacles. A trajectory extracted from the generated frames is then executed on the physical robot. The real robot follows a motion pattern consistent with the generated video, demonstrating that visually generated plans can be translated into executable robot behavior.}
  \label{fig:sim_real_frames}
\end{figure*}

Our evaluation follows the same pipeline for both robot platforms:
\begin{enumerate}
    \item A scene image is captured from a third-person perspective, showing the robot within the environment together with the target object or navigation destination.
    \item The user provides a high-level language instruction.
    \item Qwen2.5-VL-7B-Instruct refines the prompt into a visually grounded description.
    \item NVIDIA Cosmos 2.5 generates a plausible future video of the robot completing the task (Fig.~\ref{fig:sim_real_frames}).
    \item The robot is detected frame-by-frame using YOLO (Fig.~\ref{fig:pose_est}).
    \item A 6-DoF pose is estimated using a PnP solver with IPPE (Fig.~\ref{fig:pose_est}).
    \item The recovered poses are converted into a global trajectory and executed by the robot (Fig.~\ref{fig:vi_pois}).
    \item The executed trajectory is recorded using a VICON motion-capture system for comparison (Fig.~\ref{fig:exp_1},~\ref{fig:dog_results}).
\end{enumerate}

\subsection{Mobile Robot Experiments}

\begin{figure}[t]
  \centering
  \includegraphics[width=0.4\textwidth]{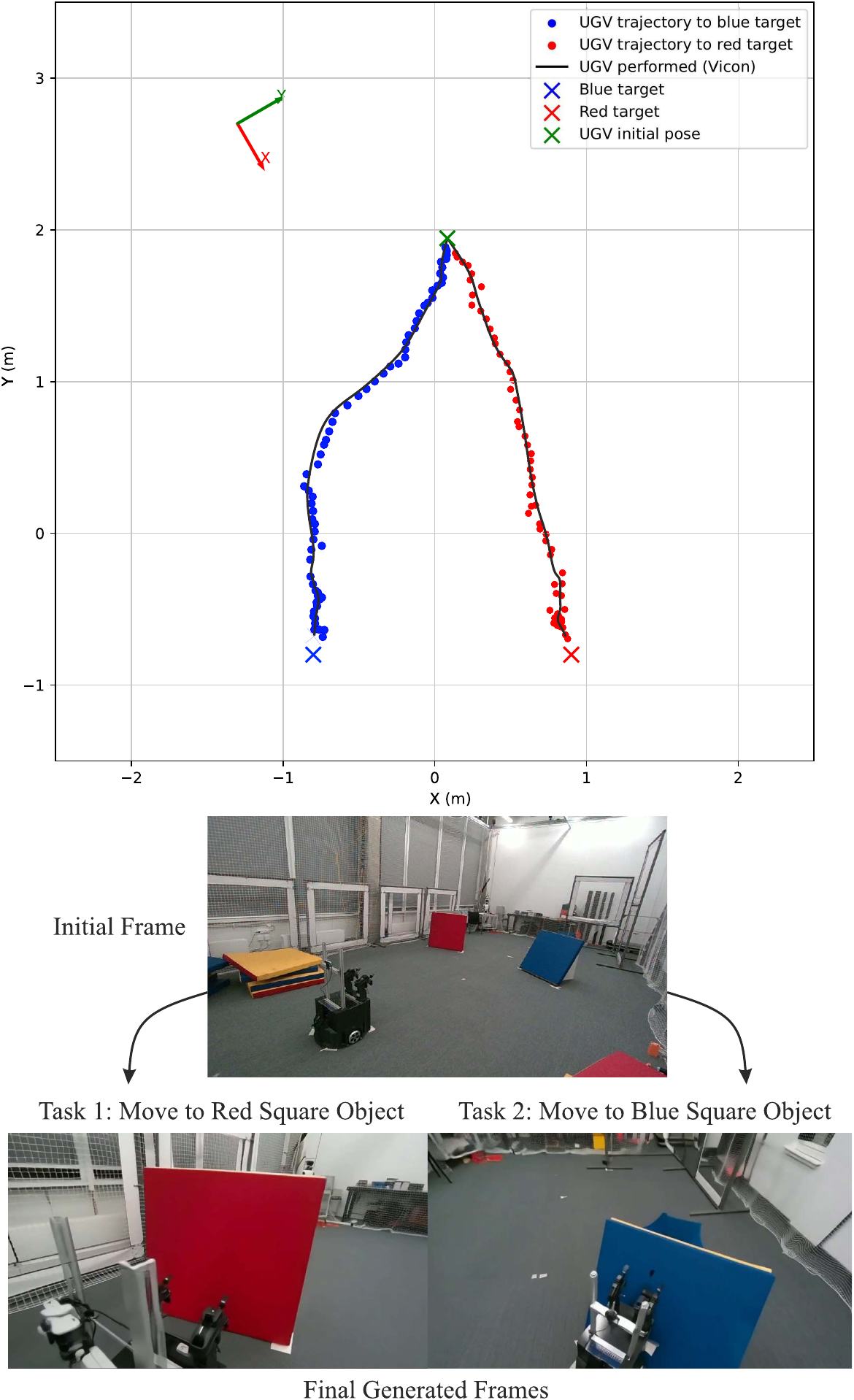}
  \caption{Experimental evaluation of the proposed navigation framework. A user first captures an initial frame of the environment. Based on this image, two navigation tasks are generated: (1) move to the red square object and (2) move to the blue square object. The figure shows the initial frame, the final generated frames corresponding to each task, and the resulting UGV trajectories. The red and blue dots represent the robot pose estimated using visual odometry during execution, while the black dashed trajectory corresponds to the ground-truth robot motion recorded by the VICON motion capture system. The axes xy indicate the camera pose of the initial captured frame.}
  \label{fig:exp_1}
\end{figure}

We first evaluate the approach on a wheeled mobile robot in an indoor arena containing colored target objects. The user specifies a target object in language, and the generated video predicts a collision-free route toward that object. Two representative tasks are shown in Fig.~\ref{fig:exp_1}: moving toward a red square target and moving toward a blue square target. The plot shows that the generated trajectories for both tasks are smooth and terminate near the correct goal objects, while the VICON-recorded real trajectories closely follow the predicted paths.

From the plotted coordinates, both UGV tasks start near $(0, 1.9)$\,m and terminate close to their designated targets. The blue-target trajectory ends near $(-0.8, -0.8)$\,m, and the red-target trajectory ends near $(0.9, -0.8)$\,m. The executed trajectories visually overlap the generated trajectories along most of the path. Based on the plot, the final position discrepancy between generated and executed trajectories is small, approximately $0.05$ - $0.10$\,m for both tasks.

\subsection{Quadruped Robot Experiments}

\begin{figure}[t]
    \centering
    \includegraphics[width=0.8\linewidth]{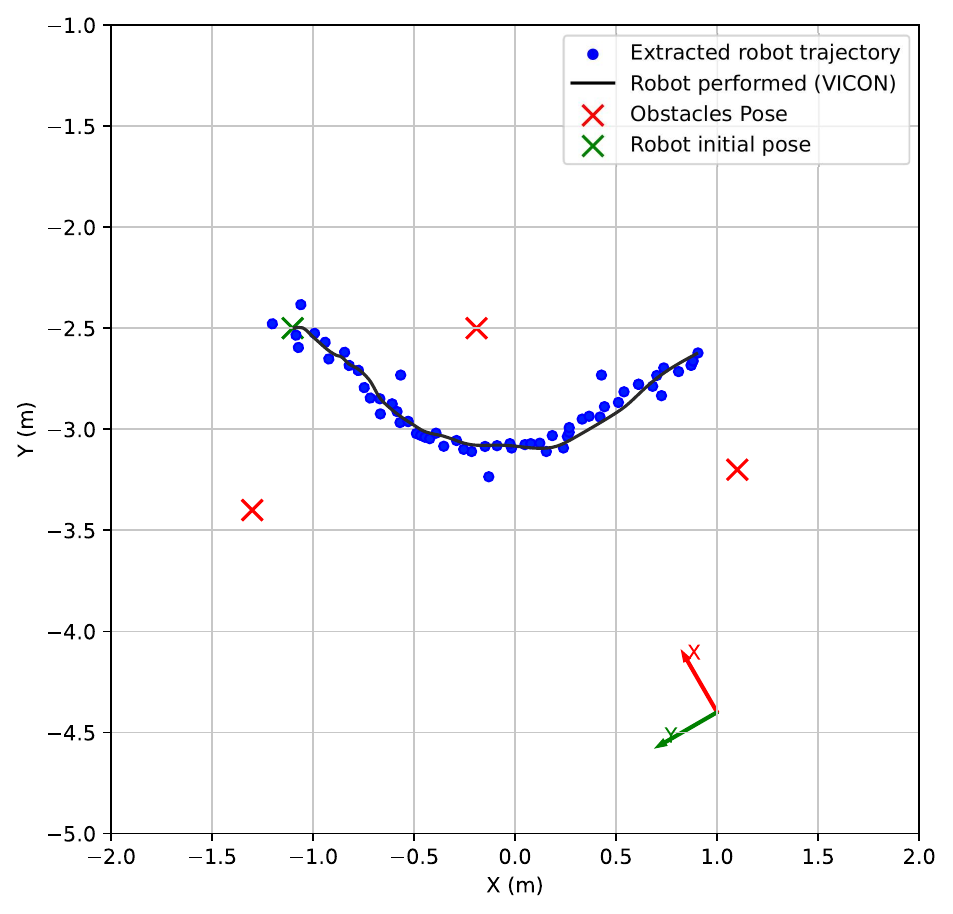}
    \caption{Quadruped obstacle-avoidance experiment. The trajectory extracted from the generated video (blue) closely follows the real trajectory recorded by VICON (black), while avoiding obstacles in the scene. The axes xy indicate the camera pose of the initial captured frame.}
    \label{fig:dog_results}
\end{figure}

We also evaluate \textit{DreamToNav} on a quadruped robot in a cluttered indoor setting containing multiple obstacles. In this case, the user instruction emphasizes safe navigation and collision avoidance. The generated video predicts a curved path that steers the robot around obstacles rather than through the center of the scene. The corresponding result, shown in Fig.~\ref{fig:dog_results} and generated and real video sequances in Fig.~\ref{fig:sim_real_frames}, compares the trajectory extracted from the generated video against the trajectory executed by the real robot and measured by VICON.

The quadruped starts near $(-1.1, -2.5)$\,m and follows a smooth arc to approximately $(0.9, -2.6)$\,m, while avoiding obstacles located near $(-1.3, -3.4)$\,m, $(-0.2, -2.5)$\,m, and $(1.1, -3.2)$\,m. The generated and executed trajectories are closely aligned along the full path. By visual inspection of the plotted error between the blue extracted points and the black VICON curve, the average tracking deviation is approximately $0.03$--$0.08$\,m, with maximum deviation remaining below roughly $0.15$\,m. The path length is approximately $2.2$--$2.4$\,m.

\section{Discussion}

Table~\ref{tab:quantitative_summary} summarizes the approximate performance values read from the reported plots. The experimental results demonstrate that DreamToNav can translate generative video predictions into executable robot trajectories across different robotic platforms. Across all experiments, the system achieved $23/30$ successful trials, corresponding to a success rate of $76.7\%$. In successful runs, the final goal error remained within $0.05$--$0.10$\,m, while trajectory tracking errors were typically below $0.15$\,m. These results indicate that the trajectories extracted from generated video sequences are sufficiently accurate for real-world navigation.

An important observation is that the proposed framework generalizes across different robot morphologies. The same pipeline was applied to both a wheeled mobile robot and a quadruped robot without modification. Despite the additional locomotion variability of the quadruped platform, both robots were able to follow the generated trajectories with comparable accuracy. Furthermore, the generated trajectories exhibit smooth, obstacle-aware motion, suggesting that the video model implicitly captures aspects of scene geometry and navigation behavior.

\begin{table}[t]
\centering
\caption{Navigation performance of DreamToNav.}
\label{tab:quantitative_summary}

\begin{tabular}{lcccc}
\hline
Task & Path (m) & Final Err (m) & Track Err (m) & Succ. \\
\hline
UGV $\rightarrow$ Red & 2.8 & 0.05--0.10 & 0.05--0.10 & 7/10 \\
UGV $\rightarrow$ Blue & 2.9 & 0.05--0.10 & 0.05--0.10 & 8/10 \\
Quadruped & 2.3 & 0.05--0.10 & 0.03--0.08 & 8/10 \\
\hline
Range & 2.3--2.9 & $<0.10$ & $<0.15$ & 23/30 \\
\hline
\end{tabular}

\end{table}

Nevertheless, the results also reveal several limitations. Failures typically occur when the generated video slightly misrepresents the scene layout or when pose estimation errors accumulate during trajectory extraction. Since the approach relies on visual generation and detection, inaccuracies in these stages can propagate into the resulting motion plan.

Overall, the experiments demonstrate that generative video planning is a promising direction for intuitive robot navigation. By allowing robots to ``imagine'' a future action sequence and then extract executable motion from it, DreamToNav provides a flexible framework for translating natural language instructions into real-world robot behavior.

\section{Conclusion}

In this work, we presented DreamToNav, a novel framework that enables robots to perform navigation tasks through generative video planning. Instead of relying on traditional geometric planners, DreamToNav interprets natural language instructions, generates a future video of the robot performing the task, and extracts an executable trajectory from the generated sequence using visual pose estimation. This approach allows robots to visually ``imagine'' actions before executing them in the real world. We evaluated the system on both a wheeled mobile robot and a quadruped robot in indoor navigation tasks. Across all experiments, the system achieved a success rate of $23/30$ trials (76.7\%), with final goal errors typically within $0.05$--$0.10$\,m and trajectory tracking errors below $0.15$\,m. These results demonstrate that trajectories derived from generative video predictions can be executed reliably across different robotic platforms.

The proposed approach highlights the potential of using generative video models as a planning mechanism for robotic control. By combining natural language understanding, visual generation, and pose-based trajectory extraction, DreamToNav provides an intuitive human-in-the-loop interface for robot navigation. Future work will focus on improving the robustness of trajectory extraction, incorporating physical constraints into the generation process, and evaluating the system in more complex environments and tasks. We believe that integrating generative models with robotic control opens a promising direction toward more flexible and intuitive robot autonomy.


\bibliographystyle{IEEEtran}
\balance
\bibliography{ref}

\end{document}